\documentclass[sigconf]{acmart}
\AtBeginDocument{%
  }

\usepackage{multirow}
\usepackage{overpic}
\usepackage[table]{xcolor}
\usepackage{balance}
\copyrightyear{2026}
\acmYear{2026}
\setcopyright{cc}
\setcctype{by}
\acmConference[MM '26] {Proceedings of the 34th ACM International Conference on Multimedia}{November 10--14, 2026}{Rio de Janeiro, Brazil}
\acmBooktitle{Proceedings of the 34th ACM International Conference on Multimedia (MM '26), November 10--14, 2026, Rio de Janeiro, Brazil}
\acmDOI{10.1145/3767308.3835247}
\acmISBN{979-8-4007-2213-4/2026/11}

\begin{document}

\title{URNet: A Unified Reparameterized Network for Efficient RGB-D Semantic Segmentation}

\author{Guoan Xu}
% \authornote{Correspondingauthor.}

\orcid{0000-0002-0181-4140}
% \author{Yang Xiao}
% \correspondingauthor
% \email{Yang.Xiao-2@student.uts.edu.au}
\affiliation{%
  \institution{University of Technology Sydney}
  \city{Sydney}
  \state{New South Wales}
  \country{Australia}
}
\email{Guoan.Xu@student.uts.edu.au}

\author{Zhengxue Wang}
\affiliation{%
  \institution{Nanjing University of Science and Technology}
  \city{Nanjing}
  \country{China}}
\email{zxwang@njust.edu.cn}

\author{Yang Xiao}
\affiliation{%
  \institution{University of Technology Sydney}
  \city{Sydney}
  \state{New South Wales}
  \country{Australia}}
\email{Yang.Xiao-2@student.uts.edu.au}

\author{Ligeng Chen}
\affiliation{%
  \institution{Honor Device Co., Ltd.}
  \city{Beijing}
  \country{China}}
\email{chenlg@smail.nju.edu.cn}

\author{Guangwei Gao}
\authornote{Corresponding author.}
\affiliation{%
  \institution{Nanjing University of Science and Technology}
  \city{Nanjing}
  \country{China}}
\email{csggao@gmail.com}

\author{Dongchen Zhu}
\affiliation{%
  \institution{Shanghai Institute of Microsystem and Information Technology, CAS}
  \city{Shanghai}
  \country{China}}
\email{dchzhu@mail.sim.ac.cn}

\renewcommand{\shortauthors}{Guoan Xu et al.}

\begin{abstract}
  Previous RGB-D semantic segmentation methods commonly employ dual encoders to separately process RGB and depth inputs, followed by dedicated modules for cross-modal feature fusion. However, such designs often inadequately capture depth representations and consequently limit effective cross-modal interaction, while the additional encoder branch introduces redundant computation that hinders lightweight execution. To tackle these challenges, we propose URNet, a Unified Reparameterized RGB-D Network that performs simultaneous multi-modal feature extraction and cross-modal fusion within a single encoder. Specifically, we adopt a reparameterization strategy to compact the network architecture and facilitate fast inference. Within each Reparameterized Block (RepBlock), a Linear Gated Attention (LGA) module is introduced to fully exploit complementary RGB and depth cues across different feature scales. Furthermore, considering that decoder design has been relatively underexplored in existing RGB-D segmentation models, we develop a concise yet effective universal decoder, termed the Pyramid Merging Decoder (PMD). Extensive experiments on multiple RGB-D segmentation benchmarks demonstrate that URNet achieves state-of-the-art performance while maintaining high efficiency. Code will be available at \url{https://github.com/Wild-Stephen/URNet}.
\end{abstract}





\maketitle
\section{Introduction}
RGB-D semantic segmentation performs pixel-wise scene parsing by jointly exploiting RGB images and depth measurements. While RGB inputs convey rich appearance cues such as color and texture, depth maps encode geometric structure and spatial relationships, serving as an essential complement under adverse conditions. 

\begin{figure}[t]
	\centering
	\includegraphics[width=0.9\linewidth]{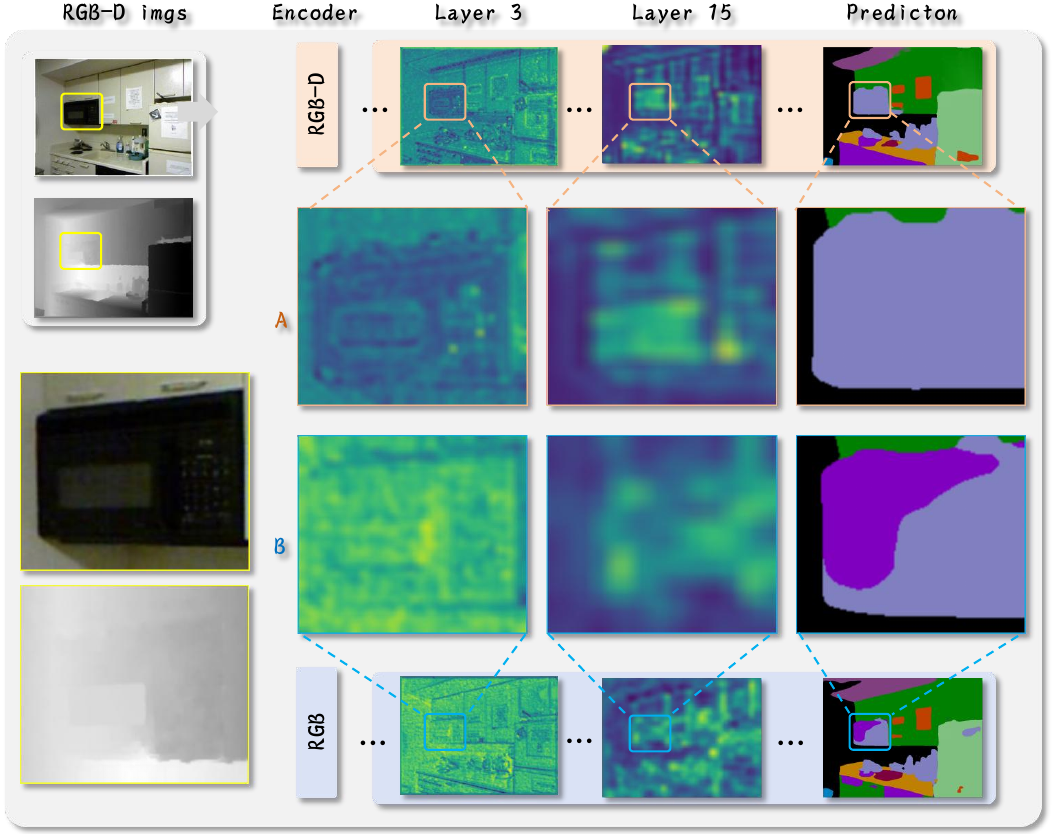}
	\caption{Feature visualization at different network stages (Layers 3, 15, and prediction) for models using RGB-D inputs but different pretraining schemes. Group A adopts a unified encoder pretrained on RGB-D, while Group B initializes both RGB and depth branches from RGB-pretrained models.} 
	\label{fig:feature}
    \Description{}
\vspace{-10pt}
\vspace{-2.5mm}
\end{figure}

Most existing approaches~\cite{xu2025adbnet,xu2026rsgmamba,wan2025sigma} adopt a dual-encoder design that decouples feature extraction and cross-modal interaction. Typically, RGB and depth features are extracted using pretrained encoders on the ImageNet dataset, after which cross-modal fusion is carried out at selected encoder stages through specialized modules. As shown in Figure~\ref{fig:feature}, although effective in practice, this framework suffers from three notable limitations: inadequate depth feature modeling due to RGB-pretrained encoders, insufficient cross-modal interaction caused by sparse fusion stages, and excessive redundancy introduced by duplicated feature extraction branches. 

% Feature visualization comparison at different network stages (Layer 3, Layer 15, and prediction) for models using RGB-D inputs but different pretraining schemes. Group A adopts a unified encoder pretrained on RGB-D data,  while in Group B, both RGB and depth branches are initialized from RGB-pretrained models.

Certain limitations have been partially mitigated in prior works~\cite{bai2025dcanet,girdhar2022omnivore,zhang2023cmx,zhang2023delivering}. DPLNet~\cite{dong2024efficient} substituted the depth encoder with multi-modal prompt generation modules, thereby reducing the computational burden of the depth branch. However, these modules are only introduced at a limited number of RGB encoder stages, which restricts the effectiveness of cross-modal fusion. GeminiFusion~\cite{jia2024geminifusion} adopted a pixel-level multi-modal fusion scheme designed to retain representations obtained from uni-modal learning, where fusion is restricted to matched tokens across modalities. Nevertheless, the limited scope of interaction constrains the effectiveness of multi-modal integration. To enable cross-modal interaction during encoding, StitchFusion~\cite{li2025stitchfusion} proposed a multi-directional Modality Adapter (MoA) that transfers information across pretrained encoders at multiple scales. While the repeated bidirectional fusion operations incur considerable computational cost, limiting the overall lightweight property and slowing down inference.

\begin{figure*}[t]
	\centering
	\includegraphics[width=0.95\textwidth]{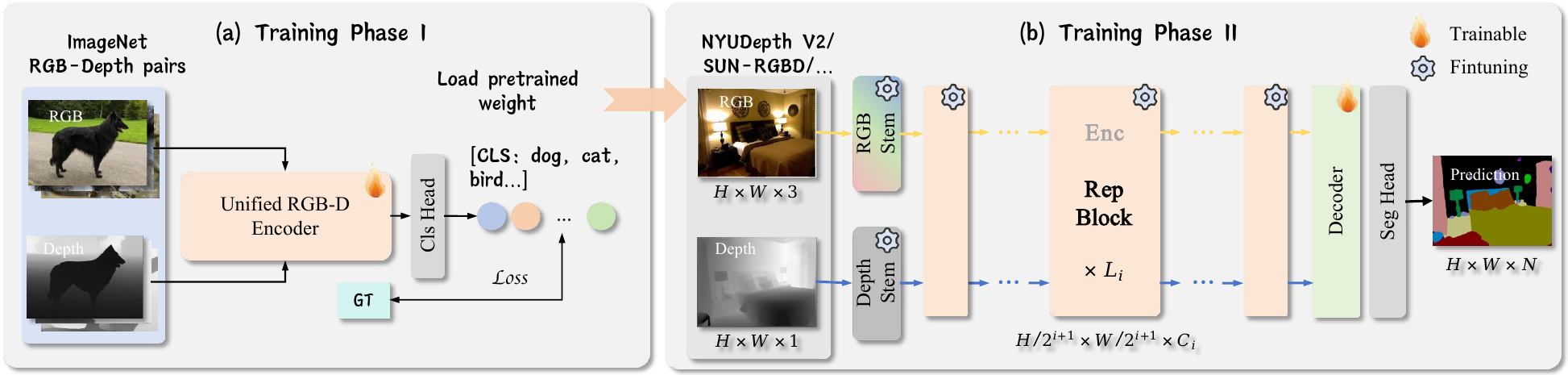}
	\caption{Overview of the training pipeline and backbone architecture of URNet.
(a) Phase-I pretraining on ImageNet-based RGB-D data.
(b) Phase-II fine-tuning on NYUDepth and so on, to adapt the pretrained representations to task-specific distributions.} 
	\label{fig:architecture}
    \Description{}
    \vspace{-2mm}
\end{figure*}

To effectively address the aforementioned limitations, we propose URNet, a lightweight and efficient RGB-D semantic segmentation framework for robotic perception. As illustrated in Figure~\ref{fig:architecture}, URNet is built upon a unified RGB-D encoder composed of multiple stacked blocks, each of which jointly performs multi-modal feature extraction and cross-modal interaction. Inspired by DFormer~\cite{yin2024dformer} and DFormer-v2~\cite{yin2025dformerv2}, we revisit the design of RGB-D pretraining and develop a dedicated pretraining framework that directly operates on paired RGB images and corresponding depth images, embedding cross-modal interactions within each encoder block. This design naturally alleviates the mismatch between pretraining and fine-tuning inputs, while enabling dense RGB–depth interaction throughout the backbone instead of restricting fusion to a limited number of stages. URNet is trained in two phases: large-scale RGB-D ImageNet-1k pretraining and RGB-D segmentation benchmark fine-tuning. Specifically, we rethink the role of reparameterization in RGB-D architectures and design a Reparameterized Block (\textbf{RepBlock}) for fast inference, in which a novel fusion mechanism, termed \textit{Linear Gated Attention (\textbf{LGA})}, is introduced to effectively integrate RGB and depth features. Furthermore, most existing networks underestimate the impact of the segmentation decoder and just use the MLP or HAM. In contrast, we introduce a lightweight \textit{Pyramid Merging Decoder (\textbf{PMD})}, which yields significant performance improvements while maintaining a slight computational burden. 

Our main contributions can be summarized as follows:
\begin{itemize}
    \item We propose URNet, a unified RGB-D pretraining framework that produces four robust RGB-D pretrained models at different scales, enabling transferable and efficient representation learning for a wide range of RGB-D downstream tasks.
    \item We introduce a lightweight yet effective architectural design, including the proposed \textbf{RepBlock} for efficient RGB–D interaction and the \textbf{Pyramid Merging Decoder (PMD)} for parameter-efficient multi-scale feature aggregation,
    \item We assess URNet on NYUDepth V2 and SUN-RGBD datasets, where it achieves competitive performance against state-of-the-art methods while maintaining a more compact architectural design.
\end{itemize}

\begin{figure*}[t]
   \centering
   %===== (a) =====%
   \begin{minipage}[b]{0.49\linewidth}
       \centering
       \includegraphics[width=\linewidth]{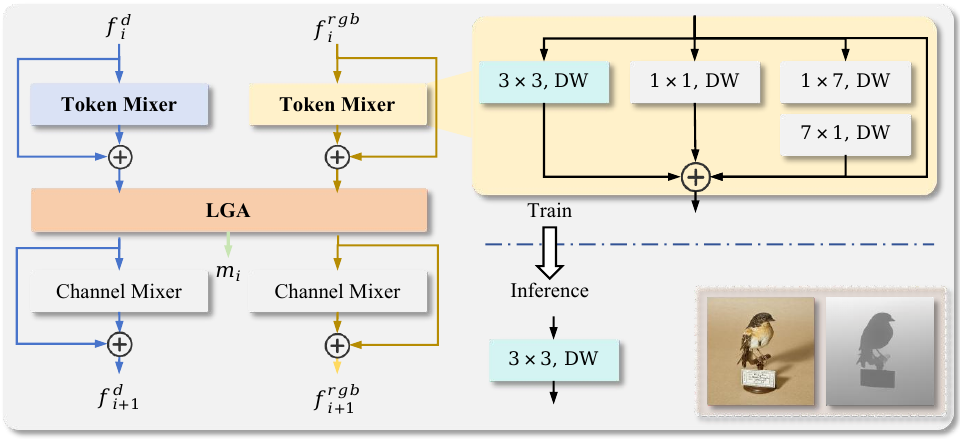}
       \\
       \footnotesize{(a) The Proposed Reparameterized block (RepBlock) architecture.}
   \end{minipage}
   \begin{minipage}[b]{0.49\linewidth}
       \centering
       \includegraphics[width=\linewidth]{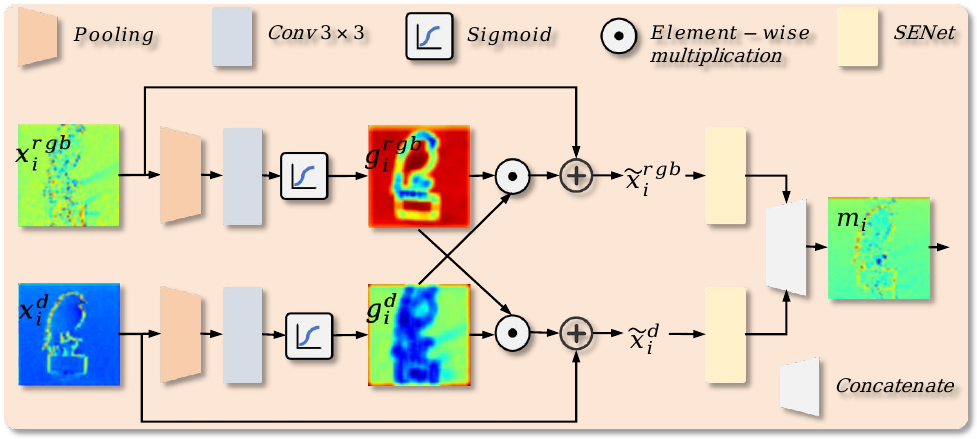}
       \\
       \footnotesize{(b) Internal design of the LGA for RGB–depth feature interaction.}
   \end{minipage}
   \caption{Detailed structure of the RepBlock with Linear Gated Attention (LGA).}
   \Description{}
   \label{fig:lga}
   \vspace{-2mm}
\end{figure*}

\section{Related Works}
\subsection{Pretrained vision backbones}

Early progress in visual representation learning was CNN backbones such as ResNets~\cite{he2016deep}, providing strong inductive biases for local and hierarchical feature modeling. More recently, Vision Transformers~\cite{dosovitskiy2020image} and hierarchical transformer backbones (e.g., Swin Transformer~\cite{liu2021swin}, PVT~\cite{wang2021pyramid}, SCASeg~\cite{xu2026scaseg} and SegFormer~\cite{xie2021segformer}) have shown strong global context modeling ability for dense prediction, albeit with increased computational and memory cost, especially at high resolutions. To improve efficiency, recent works explore hybrid and lightweight backbones that combine convolution with attention mechanisms, such as  XCiT~\cite{ali2021xcit}, S2AFormer~\cite{xu2025s2aformer}, RepViT~\cite{wang2024repvit}, and iFormer~\cite{zheng2025iformer}, which reduce token redundancy for practical deployment.

Nevertheless, when applied to multi-modal inputs, these models often suffer from feature conflicts across modalities. This issue primarily arises because they are predominantly pretrained on large-scale RGB datasets. When encountering heterogeneous modalities such as depth, the pretrained encoders tend to provide inadequate depth feature modeling, leading to suboptimal cross-modal representation learning. 

DFormer~\cite{yin2024dformer} provides an insightful perspective by leveraging paired RGB-D data to pretrain a unified RGB-D backbone. This strategy effectively alleviates the modality bias inherent in RGB-only pretraining and enables more consistent cross-modal representation learning. However, it largely overlooks the role of task-specific decoders, whose design is crucial for fully exploiting pretrained representations in downstream dense prediction tasks. Motivated by these observations, we propose a more efficient framework that further enhances computational efficiency while explicitly considering the importance of decoder design.

\subsection{RGB-D semantic segmentation}
In practical scenarios, factors such as low illumination, motion blur, and texture ambiguity often degrade the discriminative power of RGB cues, limiting their effectiveness in segmentation tasks. To mitigate these limitations, depth information has been introduced as a complementary modality, which provides explicit geometric cues about scene structure. This paradigm, commonly referred to as RGB-D semantic segmentation, leverages multi-modal sensing to enhance scene understanding. 

A common strategy in current RGB-D segmentation research is to employ dual-stream architectures and devise elaborate fusion schemes to integrate RGB and depth features learned by independently pretrained encoders. Approaches such as CMX~\cite{zhang2023cmx}, GeminiFusion~\cite{jia2024geminifusion}, AsymFormer~\cite{du2024asymformer}, and Sigma~\cite{wan2025sigma} exemplify this paradigm by adaptively merging modality-specific representations at the fusion level before the decoder. These backbones are pretrained exclusively on large-scale RGB datasets and subsequently adapted to paired RGB-D inputs during fine-tuning. This inconsistency between the pretraining setting and the multi-modal input at deployment introduces a pronounced representation distribution shift, which can undermine effective cross-modal feature alignment. Inspired by the recent advances of DFormer~\cite{yin2024dformer}, which demonstrate that pretraining a unified RGB-D model can significantly improve robustness, we followed the same principle and further aimed to reduce parameter redundancy while enhancing cross-modal feature integration, leading to a more efficient and robust RGB-D framework.

\section{Methodology}
\label {sect:method}
 Figure~\ref{fig:architecture} illustrates the pipeline of URNet, which follows an encoder–decoder architecture and is trained in two phases: RGB-D ImageNet-1k pretraining as shown in Figure~\ref{fig:architecture}(a), and fine-tuning on two RGB-D benchmarks with a lightweight decoder as depicted in Figure~\ref{fig:architecture}(b).
 % The hierarchical encoder outputs multi-scale features, ranging from high-resolution coarse features to low-resolution fine-grained representations. 

\subsection{Hierarchical Encoder}
\textbf{Overall Architecture.} The encoder of URNet adopts a multi-stage design composed of stacked RepBlocks. Specifically, these blocks are grouped into four hierarchical levels {$L_1$, $L_2$, $L_3$, $L_4$}, where $L_n$ denotes the number of blocks in the $n_{th}$ stage. In the Phase I, given an RGB image $I^{rgb}\in \mathcal{R}^{H\times W \times 3}$ and its corresponding depth map $I^{d}\in \mathcal{R}^{H\times W \times 1}$, the two modalities are initially processed independently through parallel stem layers, each composed of four $3 \times 3$ convolutional layers with a stride of 2. The features are then fed into four successive stages for thorough feature extraction and fusion, producing multi-scale feature maps at resolutions of $1/4$, $1/8$, $1/16$, and $1/32$ of the original input size. 
% Next, the encoder is pretrained on the RGB-D ImageNet-1K dataset using a classification objective to learn transferable RGB-D representations. 
In phase II, the visual features extracted by the pretrained RGB-D encoder are passed to the decoder, which generates the final predictions.
% \begin{figure}[t]
% 	\centering
% 	\includegraphics[width=1.0\linewidth]{pic/lga.pdf}
% 	\caption{LGA.} 
% 	\label{fig:lga}
%     \vspace{-2mm}
% \end{figure}

% \begin{figure*}[t]
%    \centering
%    %===== (a) =====%
%    \begin{minipage}[b]{0.49\linewidth}
%        \centering
%        \includegraphics[width=\linewidth]{pic/lga3_1.pdf}
%        \\
%        \footnotesize{(a) the Proposed Reparameterized block (RepBlock) architecture.}
%    \end{minipage}
%    \begin{minipage}[b]{0.49\linewidth}
%        \centering
%        \includegraphics[width=\linewidth]{pic/lga3_2.pdf}
%        \\
%        \footnotesize{(b) Internal design of the LGA for RGB–depth feature interaction.}
%    \end{minipage}
%    \caption{Detailed structure of the RepBlock with Linear Gated Attention (LGA).}
%    \label{fig:lga}
%    \vspace{-2mm}
% \end{figure*}

\begin{table*}[t]
\renewcommand{\arraystretch}{1.2}
\centering
\caption{Detailed configurations of the proposed URNet .}
\label{urnet}
\vspace{-3mm}
\scalebox{0.93}{
\begin{tabular}{cccccccccc}
\toprule
\multirow{2}{*}{\textbf{Stage}} & \multirow{2}{*}{\textbf{Output Size}} & \multicolumn{2}{c}{\textbf{URNet-T} }&\multicolumn{2}{c}{\textbf{URNet-S}} &\multicolumn{2}{c}{\textbf{URNet-B}} &\multicolumn{2}{c}{\textbf{URNet-L}}  \\ 
&& \textbf{RepBlocks} & \textbf{Channels} & \textbf{RepBlocks} & \textbf{Channels} & \textbf{RepBlocks} & \textbf{Channels} & \textbf{RepBlocks }&\textbf{ Channels} \\
\midrule
Stem & ($\frac{H}{4}, \frac{W}{4}$) & - &(24,24)& - &(28,28)& - &(40,40)&-  &(40,40)\\
1 & ($\frac{H}{4}, \frac{W}{4}$)&   3 & (48,48)&   5& (56,56)&   7& (80,80)&   7& (80,80)\\
2 & ($\frac{H}{8}, \frac{W}{8}$)&   4& (96,96)&   6& (112,112)&   8& (160,160)&   8& (160,160)\\
3 & ($\frac{H}{16}, \frac{W}{16}$)&   6& (192,192)&   18& (224,224)&   21& (320,320)&   36&(320,320)\\
4 & ($\frac{H}{32}, \frac{W}{32}$)&   3& (384,384)&   3& (448,448)&   3& (640,640)&   3& (640,640)\\ \midrule
\multicolumn{2}{c}{\textbf{Decoder Dim}} &\multicolumn{2}{c}{256}&\multicolumn{2}{c}{256}&\multicolumn{2}{c}{256}&\multicolumn{2}{c}{256}\\
\multicolumn{2}{c}{\textbf{Parameters (M)}} &\multicolumn{2}{c}{9.2} & \multicolumn{2}{c}{19.8} & \multicolumn{2}{c}{44.1} & \multicolumn{2}{c}{62.2}\\
% \rowcolor{gray!15}RGB-D (\textbf{ours}) & RGB-D  & 56.1\%\\
\bottomrule
\end{tabular}
}
\vspace{-2mm}
\end{table*}

\textbf{RepBlock.} Recent studies~\cite{mehta2021mobilevit} indicate that the performance of ViTs largely stems from their generic token-mixer design. Inspired by RepViT~\cite{wang2024repvit}, which leverages convolutional operations to capture global contextual information with significantly lower computational cost than self-attention, we redesign the token-mixer component using a parallel multi-branch structure. Specifically, three branches are employed in parallel: a $3 \times 3$ depthwise convolution, a $1 \times 1$ depthwise convolution, and a decomposed $1 \times 7$ followed by $7 \times 1$ convolution. Compared with RepViT~\cite{wang2024repvit}, this design effectively introduces an additional $7 \times 7$ receptive field, enabling richer spatial information aggregation.
which can be formulated as:
\begin{equation}
\begin{aligned}
x_i^{rgb} =\;& Conv^{d}_{3\times3}(f_i^{rgb}) 
            + Conv^{d}_{1\times1}(f_i^{rgb}) \\
           &+ Conv^{d}_{7\times1}\!\left(
              Conv^{d}_{1\times7}(f_i^{rgb})
             \right)
            + f_i^{rgb},
\end{aligned}
\end{equation}
where $Conv^{d}$ means depthwise convolution.
Owing to this reparameterization technique, the skip connections used during training can be removed at inference time, thereby eliminating associated computational and memory overhead. This design is particularly beneficial for deployment on resource-constrained mobile devices, as illustrated in Figure~\ref{fig:lga} (a). 

As depicted in Figure.~\ref{fig:lga} (b), in the middle of RepBlock, we introduce a Linear Gated Attention (LGA) module to perform efficient and effective fusion between RGB and depth features, as illustrated in Figure~\ref{fig:lga}. Specifically, given the RGB feature
$x_i^{rgb}$
and the corresponding depth feature $x_i^{d}$ at the $i_{th}$ stage, LGA aims to adaptively exchange and integrate complementary information between the two modalities in a lightweight manner. Specifically, both RGB and depth features are first processed independently by a pooling operation followed by a convolution layer to generate compact modality-aware descriptors. A sigmoid activation function is then applied to produce gating signals that encode the relative importance of cross-modal information. These gating signals are used to modulate the opposite modality through element-wise multiplication, enabling selective feature interaction while preserving the original modality characteristics.

Formally, the gated fusion process can be expressed as:
\begin{equation}
    g_i^{rgb}=\sigma(Conv_{3\times3}(Pool(x_i^{rgb}))),
\end{equation} 
\begin{equation}
    g_i^{d}=\sigma(Conv_{3\times3}(Pool(x_i^{d}))),
\end{equation} 
\begin{equation}
    \hat{x}_i^{rgb}=x_i^{rgb}+g_i^{d}\odot x_i^{rgb},
\end{equation} 
\begin{equation}
    \hat{x}_i^{d}=x_i^{d}+g_i^{rgb}\odot x_i^{d},
\end{equation} 
where $\sigma(\cdot)$ denotes the sigmoid function and $\odot$ represents element-wise multiplication.
After cross-gated interaction, the enhanced RGB and depth features are independently refined using lightweight Squeeze-and-Excitation (SE) modules to recalibrate channel-wise responses. The refined features are then concatenated along the channel dimension to form the final fused representation:
\begin{equation}
    m_i = Concat(SE(\hat{x}_i^{rgb}),SE(\hat{x}_i^{d})).
\end{equation}
Unlike quadratic token interactions based attention mechanisms, LGA performs fusion through linear, element-wise operations, resulting in negligible computational overhead. The symmetric cross-gating design allows RGB and depth features to mutually guide each other, enabling sufficient cross-modal interaction without introducing heavy attention blocks. This makes LGA particularly suitable for %lightweight RGB-D models and 
real-time deployment on resource-constrained devices. The channel-mixer in our model is implemented using a lightweight two-layer $1 \times 1$ convolutional structure. 

% In standard ViTs~\cite{dosovitskiy2020image}, the FFN commonly expands channel dimensions by a factor of 4, which significantly increases computation. Recent works have shown that reducing the expansion ratio can effectively alleviate this bottleneck. Specifically, LV-ViT adopts an expansion ratio of 3, and LeViT [21] further reduces it to 2. Motivated by these findings, we employ an expansion ratio of 2 for the channel mixer in all RepBlocks, achieving improved computational efficiency and faster inference.

\begin{figure}[t]
	\centering
	\includegraphics[width=0.9\linewidth]{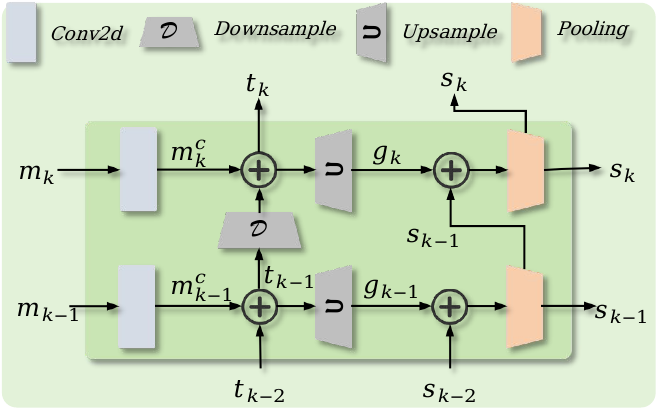}
	\caption{Structural details of the Pyramid Merging Decoder (PMD) for efficient multi-scale feature aggregation.} 
    \vspace{-3mm}
	\label{fig:decoder}
    \Description{}
    %\vspace{-1mm}
\end{figure}

\begin{table}[!t]
\centering
\caption{Quantitative results of pretrained models on rgb-d imagenet-1k trained with 4 NVIDIA H100 GPUs. GFLOPs is measured under the resolution of $224 \times224$, ‘‘\#P.'': Parameters$\downarrow$, ‘‘\#Fs.'': FLOPs$\downarrow$, and ‘‘BS'': Batch Size.}
\label{imagenet-1k}
\vspace{-3mm}
\scalebox{0.95}{
\begin{tabular}{l|ccccc}
\toprule
\textbf{Method} &\textbf{\#P.(M)$\downarrow$} & \textbf{\#Fs.(G)$\downarrow$} & \cellcolor{blue!20}\textbf{FPS$\uparrow$} & \textbf{BS}  & \textbf{Acc@1}  \\
\midrule
DFormer-T & 5.6 & 1.15 &102.2& 256 & 73.50\% \\
\rowcolor{gray!15}\textbf{URNet-T} & 9.2  & 1.35 & \textcolor{red}{\textbf{133.2}} &256 & \textbf{76.65\%}  \\
\midrule
DFormer-S & 19.0 & 3.31 &86.7&128 & 78.21\% \\
DFormerV2-S &25.4 & 4.67 &5.9&128 & 78.86\%  \\
\rowcolor{gray!15}\textbf{URNet-S} & 19.8  & 3.49 & \textcolor{red}{\textbf{105.0}} &128 & \textbf{78.95\%}  \\
\midrule
DFormer-B & 29.8 & 6.01& 59.9 &128 & 80.27\% \\
DFormer-L & 39.6 &9.88 &43.4 & 128& 82.18\% \\
DFormerV2-B &52.6 & 10.19 & 3.8 &128 & 82.32\% \\
\rowcolor{gray!15}\textbf{URNet-B} &44.1 & 8.48 & \textcolor{red}{\textbf{73.3}} & 128 & \textbf{82.49\%}  \\
\midrule
DFormerV2-L &93.8 & 19.24 &2.5 &64 &84.12\% \\
\rowcolor{gray!15}\textbf{URNet-L} &62.2 & 11.58 & \textcolor{red}{\textbf{37.0}} & 64& \textbf{84.35\%} \\
\bottomrule
\end{tabular}
}
\vspace{-15pt}
\end{table}

\subsection{Pyramid Merging Decoder (PMD)}
To effectively aggregate multi-scale features while maintaining a lightweight design, we propose a Pyramid Merging Decoder (PMD), as illustrated in Figure~\ref{fig:decoder}. Given a set of multi-scale feature maps $m_1$, $m_2$, $m_3$, and $m_4$ extracted from the encoder, where $m_1$ has the highest spatial resolution and $m_4$ has the lowest, PMD progressively merges features in a top-down pyramid manner. Specifically, for each low-level feature $m_k$ ($k = 2,3,4$), we downsample it to match the spatial resolution of its adjacent higher-level feature $m_{k-1}$ using $D(\cdot)$, and then perform element-wise addition followed by upsampling-based top-down propagation. To mitigate scale mismatch and suppress redundant information, pooling operations are applied after intermediate fusion stages. This can be formulated as
\begin{equation}
    m^c_k = Conv2d_{3\times3}(m_k),k \in \{1,2,3,4\},\\
\end{equation}
where $Conv2d_{3\times3}$ means Convolution with kernel size $3\times3$.
\begin{equation}
\begin{aligned}
t_k &= m^c_k + Ds(t_{k-1}),\\
g_k &= \mathrm{Us}\!\left(t_k\right), k \in \{2,3,4\},t_1 = g_1 = m^c_1
\end{aligned}
\end{equation}
where $Us$ represents Upsampling and $Ds$ is downsampling.
\begin{equation}
\begin{aligned}
s_k &= \mathrm{Pool}(s_{k-1} + g_k), k\in \{2,3,4\}, s_1 = Pool(g_1), \\
m_{\mathrm{f}} &= s_1 + s_2 + s_3 + s_4.
\end{aligned}
\end{equation}

\begin{table*}[t]
\begin{center}
\caption{Results on NYUDepth V2 and SUN-RGBD. All the backbones are pre-trained on ImageNet-1K.}
\label{RGBD}
\vspace{-3mm}
\scalebox{0.88}{
\begin{tabular}{lccccccccc}
\toprule
\multirow{2}{*}{\textbf{Model}} & \multirow{2}{*}{\textbf{Publication}}  & \multirow{2}{*}{\textbf{Backbone}} & \multirow{2}{*}{\textbf{\#P.(M)} $\downarrow$} & \multicolumn{3}{c}{\textbf{NYUDepth V2}} & \multicolumn{3}{c}{\textbf{SUN-RGBD}}\\ 
&&& & \textbf{Resolution} & \textbf{GFLOPs} $\downarrow$ & \textbf{mIoU} $\uparrow$ & \textbf{Resolution} & \textbf{GFLOPs} $\downarrow$ & \textbf{mIoU} $\uparrow$ \\ 
\midrule \midrule
% ACNet~\cite{hu2019acnet} & ICIP'19  & ResNet-50 & 116.6M & $480 \times 640$ & 126.7G & 48.3\% & $530 \times 730$ & 163.9G & 48.1\% \\
SGNet~\cite{chen2021spatial} & TIP'21   & ResNet-101 & 64.7 & $480 \times 640$ & 108.5 & 51.1\% & $530 \times 730$ & 151.5 & 48.6\% \\
SA-Gate~\cite{chen2020bi} & ECCV'20  & ResNet-101 & 110.9 & $480 \times 640$ & 193.7 & 52.4\% & $530 \times 730$ & 250.1 & 49.4\% \\
CEN~\cite{wang2020deep} & NeurIPS'20  & ResNet-101 & 118.2 & $480 \times 640$ & 618.7 & 51.7\% & $530 \times 730$ & 790.3 & 50.2\% \\
CEN~\cite{wang2020deep} & NeurIPS'20  & ResNet-152 & 133.9 & $480 \times 640$ & 664.4 & 52.5\% & $530 \times 730$ & 849.7 & \textcolor{red}{\textbf{51.1\%}} \\
ShapeConv~\cite{cao2021shapeconv} & CVPR'21  & ResNet-101 & 86.8 & $480 \times 640$ & 124.6 & 51.3\% & $530 \times 730$ & 161.8 & 48.6\% \\
ESANet~\cite{seichter2021efficient} & ICRA'21  & ResNet-34 & 31.2 & $480 \times 640$ & 34.9 & 50.3\% & $480 \times 640$ & 34.9 & 48.2\% \\
EMSANet~\cite{seichter2022efficient} & IJCNN'22  & ResNet-34 & 46.9 & $480 \times 640$ & 45.4 & 51.0\% & $530 \times 730$ & 58.6 & 48.4\% \\
Omnivore~\cite{girdhar2022omnivore} & CVPR'22  & Swin-T & 29.1 & $480 \times 640$ & 32.7 &  49.7\% & - & - & - \\
% Omnivore~\cite{girdhar2022omnivore} & CVPR'22  & Swin-S & 51.3M & $480 \times 640$ & 59.8G &  52.7\% & - & - & - \\
DFormer~\cite{yin2024dformer} & ICLR'24  & DFormer-S & \textcolor{blue}{\textbf{18.7}} & $480 \times 640$ & \textcolor{blue}{\textbf{25.6}} & \textcolor{red}{\textbf{53.6\%}} & $530 \times 730$ & \textcolor{blue}{\textbf{33.0}} & 50.0\%\\
% DFormer~\cite{yin2024dformer} & ICLR'24  & DFormer-T & \textcolor{red}{\textbf{6.0M}} & $480 \times 640$ & \textcolor{red}{\textbf{11.8G}} & 51.8\% & $530 \times 730$ & \textcolor{red}{\textbf{15.1G}} & 48.8\%\\
\midrule
\rowcolor{gray!15}\textbf{URNet (ours)} & 26 UR  & URNet-T & \textcolor{red}{\textbf{9.3}} & $480 \times 640$ & \textcolor{red}{\textbf{15.4}} & \textcolor{blue}{\textbf{53.2\%}} & $ 530 \times 730$ & \textcolor{red}{\textbf{19.6}} & \textcolor{blue}{\textbf{50.6\%}} \\
\midrule
% FRNet~\cite{zhou2022frnet} & JSTSP'22  & ResNet-34 & 85.5M & $480 \times 640$ & 115.6G & 53.6\% & $530 \times 730$ & 150.0G & 51.8\% \\
PGDENet~\cite{zhou2022pgdenet} & TMM'22  & ResNet-34 & 100.7 & $480 \times 640$ & 178.8 & 53.7\% & $530 \times 730$ & 229.1 & 51.0\% \\
% Omnivore~\cite{girdhar2022omnivore} & CVPR'22  & Swin-B & 95.7M & $480 \times 640$ & 109.3G &  54.0\% & - & - & - \\
TokenFusion~\cite{wang2022multimodal} & CVPR'22  & MiT-B2 & 26.0 & $480 \times 640$ & 55.2 & 53.3\% & $530 \times 730$ & 71.1 & 50.3\% \\
TokenFusion~\cite{wang2022multimodal} & CVPR'22  & MiT-B3 & 45.9 & $480 \times 640$ & 94.4 & 54.2\% & $530 \times 730$ & 122.1 & 51.0\% \\
MultiMAE~\cite{bachmann2022multimae} & ECCV'22  & ViT-B & 95.2 & $640 \times 640$ & 267.9 & \textcolor{blue}{\textbf{56.0\%}} & $640 \times 640$ & 267.9 & 51.1\% \\
CMX~\cite{zhang2023cmx} & TITS'23  & MiT-B2 &  66.6 & $480 \times 640$ & 67.6 & 54.4\% & $530 \times 730$ & 86.3 & 49.7\%\\
DCANet~\cite{bai2025dcanet} & PR'25 & VMamba & 123.8 & $480 \times 640$ & - & 53.3\% & $530 \times 730$ & - & 49.6\% \\

Sigma~\cite{wan2025sigma} & WACV'25  &VMamba-T & 48.3 & $480 \times 640$ & 90.4 & 53.9\% & $480 \times 640$ & 90.4 & 50.0\% \\
% ADBNet~\cite{xu2025adbnet} & KBS'25 & ConvNeXt & 45.9M & $480 \times 640$ & - &56.0\% & $530 \times 730$ & - & 49.6\% \\
DFormer~\cite{yin2024dformer} & ICLR'24  & DFormer-B & 29.5 & $480 \times 640$ & 41.9 & 55.6\% & $530 \times 730$ & 54.1 & \textcolor{blue}{\textbf{51.2\%}}\\
DFormerV2~\cite{yin2025dformerv2} & CVPR'25  & DFormerV2-S & \textcolor{blue}{\textbf{26.7}} & $480 \times 640$ & \textcolor{blue}{\textbf{33.9}} & \textcolor{blue}{\textbf{56.0\%}} & $530 \times 730$ & \textcolor{blue}{\textbf{43.7}} & \textcolor{blue}{\textbf{51.5\%}}\\
\midrule
\rowcolor{gray!15}\textbf{URNet (ours)} & 26 UR & URNet-S & \textcolor{red}{\textbf{19.8}} & $480 \times 640$ & \textcolor{red}{\textbf{28.5}} & \textcolor{red}{\textbf{56.1\%}} & $ 530 \times 730$ & \textcolor{red}{\textbf{36.6}} & \textcolor{red}{\textbf{51.8\%}} \\ 
\midrule
CMX~\cite{zhang2023cmx} & TITS'23  & MiT-B4 &  139.9 & $480 \times 640$ & 134.3 & 56.3\% & $530 \times 730$ & 173.8 & 52.1\%\\
CMX~\cite{zhang2023cmx} & TITS'23  & MiT-B5 &  181.1 & $480 \times 640$ & 167.8 & 56.9\% & $530 \times 730$ & 217.6 & 52.4\%\\
CMNeXt~\cite{zhang2023delivering} & CVPR'23  & MiT-B4 & 119.6 & $480 \times 640$ & 131.9 & 56.9\% & $530 \times 730$ & 170.3 & 51.9\%\\
DFormerV2~\cite{yin2025dformerv2} & CVPR'25  & DFormerV2-B & 53.9 & $480 \times 640$ & 67.2 & \textcolor{blue}{\textbf{57.7\%}} & $530 \times 730$ & 86.9 & \textcolor{blue}{\textbf{52.8\%}}\\
DFormer~\cite{yin2024dformer} & ICLR'24  & DFormer-L & \textcolor{red}{\textbf{39.0}} & $480 \times 640$ & \textcolor{blue}{\textbf{65.7}} & 57.2\% & $530 \times 730$ & \textcolor{blue}{\textbf{83.3}} & 52.5\%\\
\midrule
\rowcolor{gray!15}\textbf{URNet (ours)} & 26 UR & URNet-B & \textcolor{blue}{\textbf{43.9}} & $480 \times 640$ & \textcolor{red}{\textbf{54.0}} & \textcolor{red}{\textbf{57.9\%}} & $ 530 \times 730$ & \textcolor{red}{\textbf{73.2}} & \textcolor{red}{\textbf{53.1\%}} \\ 
\midrule
ECMRN~\cite{jia2025ecmrn} & KBS'25 & DFormer & \textcolor{blue}{\textbf{68.6}} & $480 \times 640$ & - &58.1\% & $530 \times 730$ & - & 52.9\% \\
Sigma~\cite{wan2025sigma} & WACV'25  &VMamba-S & 69.8 & $480 \times 640$ & 139.8 & 57.0\% & $480 \times 640$ & 139.8 & 52.4\% \\
DiffPixelFormer~\cite{gong2025diffpixelformer} & TMM'25 & MiT-B3 & 85.4 & $480 \times 640$ & 154.8 & 56.3\% & $530 \times 730 $ & 205.7 & 52.8\% \\
GeminiFusion~\cite{jia2024geminifusion} & ICML'24 & MiT-B3 & 75.8 & $480 \times 640$ & - & 56.8\% & $530 \times 730$ & - & 52.7\% \\
DFormerV2~\cite{yin2025dformerv2} & CVPR'25  & DFormerV2-L & 95.5 & $480 \times 640$ & \textcolor{blue}{\textbf{124.1}} & \textcolor{blue}{\textbf{58.4\%}} & $530 \times 730$ & \textcolor{blue}{\textbf{160.5}} & \textcolor{blue}{\textbf{53.3\%}}\\
\midrule
\rowcolor{gray!15}\textbf{URNet (ours)} & 26 UR & URNet-L & \textcolor{red}{\textbf{61.8}} & $480 \times 640$ & \textcolor{red}{\textbf{73.0}} & \textcolor{red}{\textbf{58.8\%}} & $ 530 \times 730$ & \textcolor{red}{\textbf{94.6}} & \textcolor{red}{\textbf{53.4\%}} \\ 
\bottomrule
\end{tabular}
}
\end{center}
\end{table*}

This progressive merging strategy enables effective information propagation across scales while preserving computational efficiency. After multi-scale feature aggregation, a pyramid pooling (P-Pooling) module is applied to capture contextual information at different receptive fields. Finally, the pooled features are fused with the aggregated representation via a residual connection, producing the final decoder output. This can be formulated as
\begin{equation}
    y=m_{f}+PPool(m_{f}),
\end{equation}
where $PPool$ means pyramid pooling. Owing to its simple operations (upsampling, pooling, and element-wise addition), PMD introduces minimal parameter overhead while significantly enhancing segmentation performance.

\section{Experiments}
\label {sect:experiment}
\subsection{Datasets and implementation details}
Consistent with prior RGB-D segmentation studies~\cite{yin2024dformer},~\cite{yin2025dformerv2}, URNet is pretrained on RGB-D ImageNet-1k and then fine-tuned on NYUDepth V2 and SUN-RGBD. NYUDepth V2 contains 795/654 training/testing images with 40 classes, while SUN-RGBD includes 10,335 samples across 37 categories (5,285/5,050 split). Training is conducted on 4 NVIDIA H100 GPUs using a polynomial learning rate schedule with 300 epochs. During fine-tuning, we apply random flipping and scale ranging (0.5–1.75) as data augmentations. Images are resized to $480 \times 640$ for NYUDepth V2 and $480 \times 480$ for SUN-RGBD. The detailed architectural configurations for different model sizes of URNet are presented in Table~\ref{urnet}.

\begin{figure*}[t]
	\centering
	\begin{overpic}[width=0.9\textwidth]{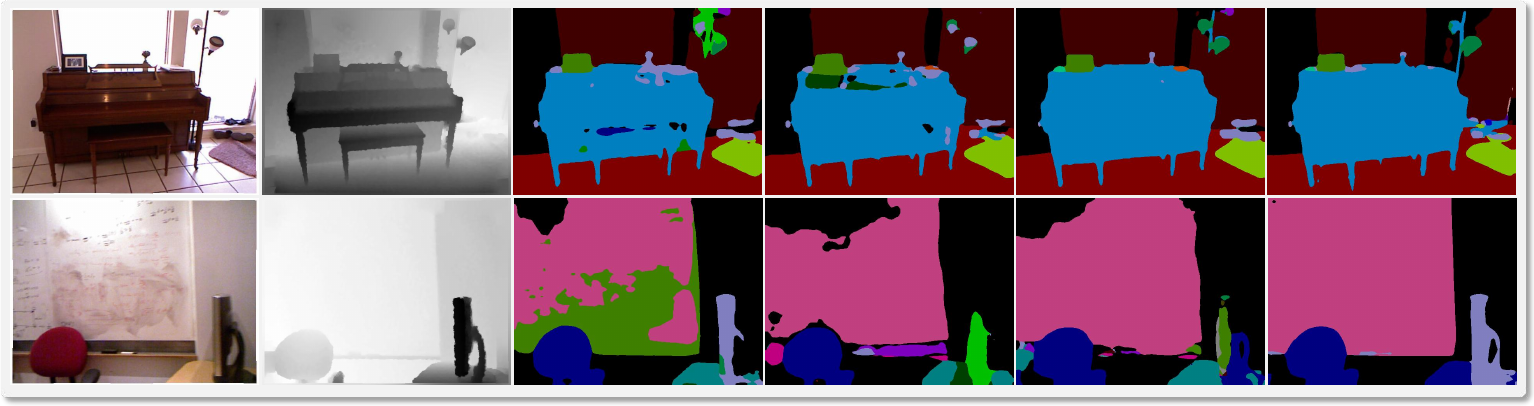}
        \put(6,-1.5){\small RGB Imgs}
        \put(22,-1.5){\small Depth Imgs}
        \put(39,-1.5){\small CMNeXt}
        \put(55,-1.5){\small DFormer-B}
        \put(69.5,-1.5){\small DFormerV2-B}
        \put(85,-1.5){\small \textbf{URNet-B (ours)}}
    \end{overpic}
	\caption{Qualitative comparisons on NYU-Depth V2. From left to right are RGB images, depth images, and segmentation results produced by CMNeXt~\protect\cite{zhang2023delivering}, DFormer-B~\protect\cite{yin2024dformer}, DFormerV2-B~\protect\cite{yin2025dformerv2}, and URNet-B (ours).}
	\label{nyu_vis}
    \Description{}
    \vspace{-3mm}
\end{figure*}

\begin{figure*}[t]
	\centering
	\begin{overpic}[width=0.9\textwidth]{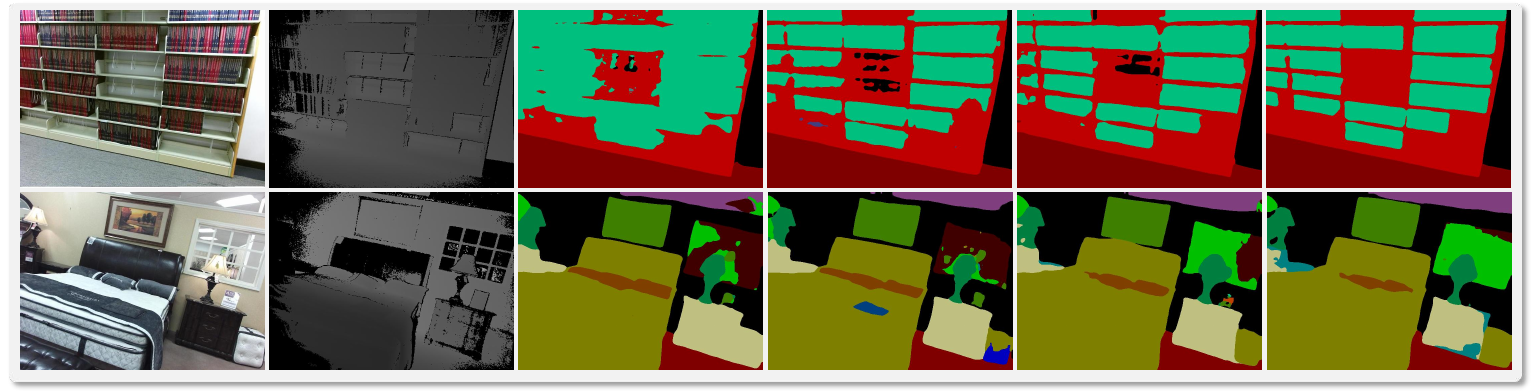}
        \put(6,-1.5){\small RGB Imgs}
        \put(22,-1.5){\small Depth Imgs}
        \put(39,-1.5){\small CMNeXt}
        \put(55,-1.5){\small DFormer-B}
        \put(69.5,-1.5){\small DFormerV2-B}
        \put(85,-1.5){\small \textbf{URNet-B (ours)}}
    \end{overpic}
	\caption{Qualitative comparisons on SUN RGB-D. From left to right are RGB images, depth images, and segmentation results produced by CMNeXt~\protect\cite{zhang2023delivering}, DFormer-B~\protect\cite{yin2024dformer}, DFormerV2-B~\protect\cite{yin2025dformerv2}, and URNet-B (ours).}
	\label{sun_vis}
    \Description{}
    \vspace{-3mm}
\end{figure*}

\subsection{Comparison with state-of-the-art methods}
Table~\ref{imagenet-1k} presents the results of the training phase I, i.e., pretraining on the RGB-Depth ImageNet-1K dataset. It reports the quantitative performance of different RGB-D pretrained backbones. All models are trained under the same RGB-D setting using 4 NVIDIA H100 GPUs. Under similar model scales, URNet consistently achieves a better accuracy–efficiency trade-off than that of DFormer. For example, at the Tiny scale, URNet-T improves the Top-1 accuracy by +3.15\% (76.65\% vs. 73.50\%) compared with DFormer-T, while incurring only a moderate increase in parameters and FLOPs. Notably, URNet-T also runs faster in inference, achieving 133.2 FPS versus 102.2 FPS. Similar trends can be observed at the Small and Base scales. URNet-S slightly outperforms DFormer-S in Top-1 accuracy (78.95\% vs. 78.21\%) while maintaining comparable model complexity and significantly higher inference speed (105.0 FPS vs. 86.7 FPS). URNet-B further improves accuracy to 82.49\%, surpassing both DFormer-B and DFormerV2-B, while achieving higher throughput. Compared with the stronger DFormerV2 variants, URNet-B achieves a similar Top-1 accuracy to DFormerV2-B (82.49\% vs. 82.32\%) while being over $19\times$ faster in inference (73.3 FPS vs. 3.8 FPS). This demonstrates that URNet offers a more efficient solution for large-scale RGB-D pretraining, especially in scenarios with strict efficiency constraints. 
% At the Large scale, URNet-L achieves strong performance with 84.35\% Top-1 accuracy, slightly surpassing DFormerV2-L, while maintaining a dramatically higher inference speed (37.0 FPS vs. 2.5 FPS), avoiding the excessive computational cost and latency introduced by DFormerV2-L.

During training phase II, we conduct comparisons between URNet and a set of state-of-the-art and representative RGB-D semantic segmentation methods on the NYUDepthv2 and SUN-RGBD benchmarks. Quantitative results are summarized in Table~\ref{RGBD}. Early CNN-based methods built on ResNet backbones generally require heavy computation to achieve competitive performance, with FLOPs often exceeding 100G while yielding limited accuracy gains on both NYU Depth V2 and SUN-RGBD. Compared with recent DFormer-based approaches, URNet consistently gets a better accuracy–efficiency trade-off across different model scales. At the Tiny scale, URNet-T achieves comparable performance to DFormer-S while using significantly fewer parameters and FLOPs, reaching 53.2\% mIoU on NYU Depth V2 and 50.6\% on SUN-RGBD with only 15.4G and 19.6G FLOPs, respectively. At the Small scale, URNet-S slightly surpasses DFormerV2-S in accuracy on both datasets (56.1\% vs. 56.0\% on NYU Depth V2, and 51.8\% vs. 51.5\% on SUN-RGBD), while requiring fewer parameters and around 20\% less computation. At the Base scale, URNet-B further improves the performance to 57.9\% mIoU on NYU Depth V2 and 53.1\% on SUN-RGBD, outperforming both DFormer-B and DFormerV2-B with notably lower FLOPs. At the Large scale, URNet-L achieves strong performance with 58.8\% mIoU on NYU Depth V2 and 53.4\% on SUN-RGBD. Although DFormerV2-L attains slightly lower accuracy on SUN-RGBD and comparable performance on NYU Depth V2, it requires substantially more computation (124.1G vs. 73.0G FLOPs), indicating that URNet-L preserves strong representation capability while avoiding the excessive computational overhead of large-scale DFormerV2 models. Overall, these results demonstrate that URNet consistently delivers competitive or superior performance under significantly reduced computational budgets, highlighting its effectiveness as an efficient RGB-D semantic segmentation backbone. The qualitative visualization results are illustrated in Figure~\ref{nyu_vis} and Figure~\ref{sun_vis}.

\subsection{Ablation Studies}
\textbf{RGB vs. RGBD pretraining.} As shown in Table~\ref{abl_pretrain}, under identical finetuning settings, RGB–D pretraining achieves 56.1\% mIoU, outperforming RGB-only pretraining (54.9\% mIoU) by +1.2\%. This gain demonstrates that incorporating depth information during pretraining facilitates better cross-modal feature alignment and improves downstream RGB–D segmentation performance.

\begin{table}[t]
\centering
\caption{Comparisons between RGB vs. RGB-D pretraining under the same baseline URNet-S for NYUDepth V2.}
\label{abl_pretrain}
\vspace{-3mm}
\scalebox{0.9}{
\begin{tabular}{lcc}
\toprule
\textbf{Pretrain} & \textbf{Finetune}  & \textbf{mIoU(\%)$\uparrow$}
\\ \midrule
RGB-RGB & RGB-D & 54.9 \\
\rowcolor{gray!15}RGB-D (\textbf{ours}) & RGB-D  & 56.1\\
\bottomrule
\end{tabular}
}
\vspace{-2mm}
\end{table}

\begin{table}[t]
\centering
\caption{Ablation studies of different token mixers. SA refers to self-attention, XCiT~\protect\cite{ali2021xcit} denotes cross-covariance image transformers, Conv represents convolutional operators, and Rep indicates reparameterization.}
\label{abl_tokenmix}
\vspace{-3mm}
\scalebox{0.9}{
\begin{tabular}{lcccc}
\toprule
\textbf{Token Mixer} & \textbf{\#P.(M)$\downarrow$} &\textbf{\#Fs.(G)$\downarrow$} & \textbf{FPS$\uparrow$} & \textbf{mIoU(\%)$\uparrow$} \\ \midrule
SA   & 18.40 & 20.89 & 18.23 & 53.45\\
XCiT  & 14.27 & 21.49 & 94.87 & 52.61\\
Conv   & 10.20 & 16.27 & 126.84 & 51.35\\
\rowcolor{gray!15}\textbf{Rep. (ours)} & \textbf{9.26} & \textbf{15.35} & 100.35 & 53.17  
\\
\bottomrule
\end{tabular}
}
\vspace{-2mm}
\end{table}

\begin{table}[t]
\centering
\caption{Ablation studies of Linear Gated Attention (LGA).}
\label{abl_lga}
\vspace{-3mm}
\scalebox{0.9}{
\begin{tabular}{ccccccc}
\toprule
\multicolumn{3}{c}{\textbf{Fusion}} & \multirow{2}{*}{\textbf{\#P.(M)$\downarrow$}} &\multirow{2}{*}{\textbf{\#Fs.(G)$\downarrow$}} &\multirow{2}{*}{\textbf{FPS$\uparrow$}} &\multirow{2}{*}{\textbf{mIoU(\%)$\uparrow$}} \\
\textbf{Add} & \textbf{Concat} & \textbf{LGA} & \\ \midrule
\checkmark &  && 8.55 & 15.35 & 129.09 & 51.83\\
&   \checkmark && 9.26 & 16.06 & 87.79 & 52.32\\
\rowcolor{gray!15}&  &  \checkmark & 9.26 & 15.35 & 100.35 & \textbf{53.17}\\

\bottomrule
\end{tabular}
}
\vspace{-2mm}
\end{table}

\begin{table}[!t]
\centering
\caption{Ablation studies of Pyramid Merging Decoder (PMD), MLP: Multi-Layer Perception, HAM: Hamburger.}
\label{abl_pmd}
\vspace{-3mm}
\scalebox{0.9}{
\begin{tabular}{ccccccc}
\toprule
 \multicolumn{3}{c}{\textbf{Decoder}}  & \multirow{2}{*}{\textbf{\#P.(M) $\downarrow$}} & \multirow{2}{*}{\textbf{\#Fs.(G) $\downarrow$}} &\multirow{2}{*}{\textbf{FPS $\uparrow$}}
& \multirow{2}{*}{\textbf{mIoU(\%) $\uparrow$}}\\
\textbf{MLP} & \textbf{HAM} & \textbf{PMD} \\
\midrule
 \checkmark & & &8.93 &13.98 & 120.51 & 51.74 \\
& \checkmark  &&9.52&18.40 & 105.40 & 52.78  \\
\rowcolor{gray!15}& & \checkmark &9.26 &15.35&100.35&\textbf{53.17} \\

\bottomrule
\end{tabular}
}
\vspace{-3mm}
\end{table}

% \begin{table}[!t]
% \centering
% \caption{Comparisons between RGB vs. RGB-D pretraining under the same baseline URNet-S for the NYUDepth V2 dataset.}
% \label{abl_pretrain}
% \scalebox{1.0}{
% \begin{tabular}{lcc}
% \toprule
% \textbf{Pretrain} & \textbf{Finetune}  & \textbf{mIoU(\%)$\uparrow$}
% \\ \midrule
% RGB-RGB & RGB-D & 54.9 \\
% \rowcolor{gray!15}RGB-D (\textbf{ours}) & RGB-D  & 56.1\\
% \bottomrule
% \end{tabular}
% }
% \vspace{-3mm}
% \end{table}

\textbf{Effectiveness of Token Mixer.} We conduct an ablation study of different token mixers on the NYUDepth V2 dataset, as shown in Table~\ref{abl_tokenmix}. Self-attention (SA) achieves the highest mIoU (53.45\%), but incurs heavy computational overhead (20.89G FLOPs) and low inference speed (18.23 FPS). In contrast, XCiT and convolution-based mixers significantly reduce parameters and improve throughput, but suffer from clear performance drops, yielding 52.61\% and 51.35\% mIoU, respectively. The proposed $Rep.$ token mixer provides a more favorable accuracy–efficiency trade-off. Compared with SA, $Rep.$ attains comparable accuracy (53.17\% vs. 53.45\%) while reducing parameters from 18.40M to 9.26M and FLOPs from 20.89G to 15.35G, and improving inference speed by over $5\times$ (100.35 vs. 18.23 FPS). Compared with XCiT and convolution-based mixers, $Rep.$ achieves notably higher mIoU with competitive computational cost, demonstrating its effectiveness for efficient RGB-D semantic segmentation on NYUDepth V2.

\begin{figure}[t]
	\centering
	\begin{overpic}[width=0.45\textwidth]{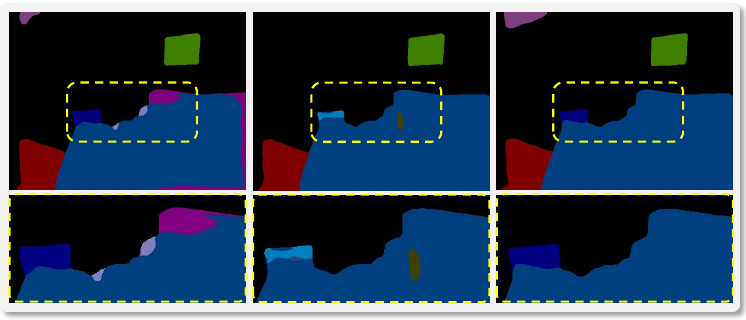}
        \put(10,-3){\small (a) Add}
        \put(45,-3){\small (b) Concat}
        \put(78,-3){\small (\textbf{c) LGA}}
        % \put(24,1){\small Depth Imgs}
        % \put(40.5,1){\small CMNeXt}
        % \put(56,1){\small DFormer-B}
        % \put(70.5,1){\small DFormerV2-B}
        % \put(86,1){\small \textbf{URNet-B (ours)}}
    \end{overpic}
    \vspace{-1mm}
	\caption{Visualizations under different fusion strategies.}
	\label{fig:vs_lga}
    \Description{}
    \vspace{-3mm}
\end{figure}

\textbf{Effectiveness of LGA.} Table~\ref{abl_lga} reports the ablation study of different fusion strategies for the proposed Linear Gated Attention (LGA) on the NYUDepth V2 dataset. Using simple element-wise addition results in the lowest mIoU, indicating that naive fusion is insufficient to effectively exploit the complementary information between RGB and depth modalities. Replacing addition with channel concatenation brings a modest improvement (from 51.83\% to 52.32\% mIoU), but at the cost of increased parameters and FLOPs. In contrast, the proposed LGA achieves a substantial performance gain on NYUDepth V2, reaching 53.17\% mIoU, while even reducing computational cost compared with concatenation. This demonstrates that LGA enables more effective and efficient cross-modal interaction by dynamically re-weighting features from different modalities. For better comparison, the visualizations under different fusion methods are shown in Figure~\ref{fig:vs_lga}.

\begin{figure}[t]
	\centering
	\begin{overpic}[width=0.45\textwidth]{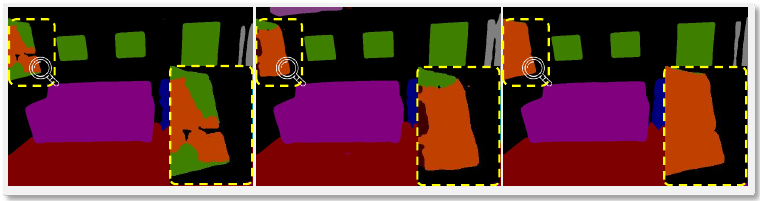}
        \put(10,-3){\small (a) MLP}
        \put(45,-3){\small (b) HAM}
        \put(78,-3){\small (\textbf{c) PMD}}
        % \put(24,1){\small Depth Imgs}
        % \put(40.5,1){\small CMNeXt}
        % \put(56,1){\small DFormer-B}
        % \put(70.5,1){\small DFormerV2-B}
        % \put(86,1){\small \textbf{URNet-B (ours)}}
    \end{overpic}
    \vspace{-1mm}
	\caption{Qualitative comparison under different decoders.}
	\label{fig:vs_dec}
    \Description{}
    \vspace{-3mm}
\end{figure}

\begin{table}[!t]
\centering
\caption{Comparisons with Different Encoder–Decoders.}
\label{abl_backbone}
\vspace{-3mm}
\scalebox{0.9}{
\begin{tabular}{lccc}
\toprule
\textbf{Encoder}   & \textbf{\#P.(M)$\downarrow$} & \textbf{Decoder}& \textbf{mIoU(\%)$\uparrow$} \\ \midrule

Sigma-T~\cite{wan2025sigma} & 45.7 &MLP & 52.95\\
\rowcolor{gray!15}URNet-B (\textbf{ours}) & 40.5 & MLP & 56.74\\
\midrule
Sigma-T~\cite{wan2025sigma} & 48.1 &PMD & 54.63\\
\rowcolor{gray!15}URNet-B (\textbf{ours}) & 43.9 & PMD & 57.91\\
\midrule\midrule
DFormerV2-L~\cite{yin2025dformerv2} & 95.5 & HAM & 58.38\\
\rowcolor{gray!15}URNet-L (\textbf{ours}) & 72.6 & HAM & 58.22\\
\midrule
DFormerV2-L~\cite{yin2025dformerv2} &84.6 & PMD & 58.66\\
\rowcolor{gray!15}URNet-L (\textbf{ours}) & 61.8& PMD & 58.75\\

\bottomrule
\end{tabular}
}
\vspace{-3mm}
\end{table}

\begin{table*}[t]
\centering
\caption{Quantitative comparisons on RGB-D SOD benchmarks. 
We use the mean absolute error ($M\downarrow$), max F-measure ($F_m\uparrow$), S-measure ($S_m\uparrow$), and max E-measure ($E_m\uparrow$) as evaluation metrics. SPNet~\protect\cite{zhou2021specificity}, VST~\protect\cite{liu2021visual}, RD3D+~\protect\cite{chen20223}, SPSN~\protect\cite{lee2022spsn}, DFormer~\protect\cite{yin2024dformer}, and DFormerV2~\protect\cite{yin2025dformerv2}.}
\label{SOD}
\vspace{-3mm}
\scalebox{0.75}{
\begin{tabular}{lcc|cccc|cccc|cccc|cccc|cccc}
\toprule
\multirow{2}{*}{\textbf{Method}} 
& \multicolumn{2}{c|}{\textbf{Datasets}} 
& \multicolumn{4}{c|}{\textbf{DES}} 
& \multicolumn{4}{c|}{\textbf{NLPR}} 
& \multicolumn{4}{c|}{\textbf{NJU2K}} 
& \multicolumn{4}{c|}{\textbf{STERE}} 
& \multicolumn{4}{c}{\textbf{SIP}} \\
\cmidrule(lr){2-3}
\cmidrule(lr){4-7} \cmidrule(lr){8-11} \cmidrule(lr){12-15} \cmidrule(lr){16-19} \cmidrule(lr){20-23}
& \textbf{\#P.(M)} & \textbf{\#F.(G)}
& \textbf{$M$} & \textbf{$F_m$} & \textbf{$S_m$} & \textbf{$E_m$}
& \textbf{$M$} & \textbf{$F_m$} & \textbf{$S_m$} & \textbf{$E_m$}
& \textbf{$M$} & \textbf{$F_m$} & \textbf{$S_m$} & \textbf{$E_m$}
& \textbf{$M$} & \textbf{$F_m$} & \textbf{$S_m$} & \textbf{$E_m$}
& \textbf{$M$} & \textbf{$F_m$} & \textbf{$S_m$} & \textbf{$E_m$} \\
\midrule
SPNet & 150.3 &68.1 &.014 &.950 &.945 &.980 &.021 &.925 &.927 &.959 &.028 &.935 &.925 &.954 &.037 &.915 &.907 &.944 &.043 &.916 &.894 &.930 \\
VST & 83.3 &31.0 &.017 &.940 &.943 &.978 &.024 &.920 &.932 &.962 &.035 &.920 &.922 &.951 &.038 &.907 &.913 &.951 &.040 &.915 &.904 &.944 \\
RD3D+ & 28.9 &43.3 &.017 &.946 &.950 &.982 &.022 &.921 &.933 &.964 &.033 &.928 &.928 &.955 &.037 &.905 &.914 &.946 &.046 &.900 &.892 &.928 \\
SPSN & 37.0 & 100.3 & .017 &.942 &.937 &.973 &.023 &.917 &.923 &.956 &.032 &.927 &.918 &.949 &.035 &.909 &.906 &.941 &.043 &.910 &.891 &.932\\
HiDANet & 130.6 &71.5 &.013 &.952 &.946 &.980 &.021 &.929 &.930 &.961 &.029 &.939 &.926 &.954 &.035 &.921 &.911 &.946 &.043 &.919 &.892 &.927 \\
\midrule
DFormer-T &\textbf{5.9} &4.5 &.016 &\textcolor{red}{\textbf{.947}} &.941 &.975 &\textcolor{red}{\textbf{.021}} &.931 &.932 &.960 &\textcolor{red}{\textbf{.028}} &.937 &.927 &.953 &\textcolor{red}{\textbf{.033}} &.921 &.915 &\textcolor{red}{\textbf{.945}} &.039 &\textcolor{red}{\textbf{.922}} &.900 &\textcolor{red}{\textbf{.935}} \\
\rowcolor{gray!15}\textbf{URNet-T} & 8.6 & \textbf{2.6} & \textcolor{red}{\textbf{.015}} &.944 & \textcolor{red}{\textbf{.950}} & \textcolor{red}{\textbf{.979}} & \textcolor{red}{\textbf{.021}} & \textcolor{red}{\textbf{.932}} & \textcolor{red}{\textbf{.933}} & \textcolor{red}{\textbf{.960}} & .029 & \textcolor{red}{\textbf{.939}} & \textcolor{red}{\textbf{.928}} & \textcolor{red}{\textbf{.957}} & .035 &\textcolor{red}{\textbf{.925}} & \textcolor{red}{\textbf{.917}} & .944 & \textcolor{red}{\textbf{.038}} & .920 & \textcolor{red}{\textbf{.901}} & .932\\
\midrule
DFormer-S &\textbf{18.5} &10.1 &.016 &\textcolor{red}{\textbf{.950}} &.939 &.970 &\textcolor{red}{\textbf{.020}} &\textcolor{red}{\textbf{.937}} &.936 &.965 &.026 &.941 &.931 &.960 &.031 &.928 &.920 &\textcolor{red}{\textbf{.951}} &.041 &.921 &.898 &.931 \\
DFormerV2-S &-&- &.014 &.946 & .951 & .980 &.021 & .932 &.938 & \textcolor{red}{\textbf{.967}} & \textcolor{red}{\textbf{.025}}& .933 & .937 & .958 & .031 & .926 & \textcolor{red}{\textbf{.921}} & .950  & .038 & .922 & \textcolor{red}{\textbf{.912}} & .935\\
\rowcolor{gray!15}\textbf{URNet-S} & 19.1 & \textbf{6.7} & \textcolor{red}{\textbf{.013}} & .947 & \textcolor{red}{\textbf{.952}} & \textcolor{red}{\textbf{.982}} & .021 & \textcolor{red}{\textbf{.937}} & \textcolor{red}{\textbf{.939}} & .966 & .026 & \textcolor{red}{\textbf{.947}} & \textcolor{red}{\textbf{.938}} & \textcolor{red}{\textbf{.965}} & \textcolor{red}{\textbf{.030}} & \textcolor{red}{\textbf{.933}} & \textcolor{red}{\textbf{.921}} & .946 & \textcolor{red}{\textbf{.037}} & \textcolor{red}{\textbf{.923}} & .908 & \textcolor{red}{\textbf{.937}}\\
\midrule
DFormer-B &\textbf{29.3} &16.7 &\textcolor{red}{\textbf{.013}} &.957 &.948 &.982 &.019 &.933 &.936 &.965 &.025 &.941 &.933 &.960 &.029 &\textcolor{red}{\textbf{.931}} &.925 &.951 &.035 &\textcolor{red}{\textbf{.932}} &.908 &.943 \\
DFormerV2-B &-&-&.017 & .937 & .943 & .969 & \textcolor{red}{\textbf{.017}} & .936 & \textcolor{red}{\textbf{.939}} &.966 & \textcolor{red}{\textbf{.024}} & .938 & \textcolor{red}{\textbf{.950}} & .963 & .030 & .925 & .926 & \textcolor{red}{\textbf{.952}} & .039 & .902 & \textcolor{red}{\textbf{.931}} & .940\\
\rowcolor{gray!15}\textbf{URNet-B} & 43.2 & \textbf{16.3} & \textcolor{red}{\textbf{.013}} & \textcolor{red}{\textbf{.963}} & \textcolor{red}{\textbf{.950}} & \textcolor{red}{\textbf{.984}} & \textcolor{red}{\textbf{.017}} & \textcolor{red}{\textbf{.940}} & .928 & \textcolor{red}{\textbf{.967}} & \textcolor{red}{\textbf{.024}} & \textcolor{red}{\textbf{.943}} & .931 &\textcolor{red}{\textbf{.968}} & \textcolor{red}{\textbf{.027}} & .930 & \textcolor{red}{\textbf{.927}} & \textcolor{red}{\textbf{.952}} & \textcolor{red}{\textbf{.034}} & .930 & .912 & \textcolor{red}{\textbf{.945}}\\
\midrule
DFormer-L &\textbf{38.8} &26.2 &.013 &.956 &.948 &.980 &.016 &.939 &.942 &.971 &\textcolor{red}{\textbf{.023}} &.946 &.937 &.964 &.030 &\textcolor{red}{\textbf{.929}} &.923 &.952 &\textcolor{red}{\textbf{.032}} &.938 &.915 &.950 \\
DFormerV2-L &-&-& .017 & .940 & .945 & .973 & .016 & \textcolor{red}{\textbf{.941}} & .943 & .969 & .024 & .935 & \textcolor{red}{\textbf{.945}} & .961 & .029 & .925 & .928 & .952 & .038 & .901 & \textcolor{red}{\textbf{.928}} & .936\\
\rowcolor{gray!15}\textbf{URNet-L} & 62.0 & \textbf{22.3}&\textcolor{red}{\textbf{.012}} & \textcolor{red}{\textbf{.962}} & \textcolor{red}{\textbf{.954}} & \textcolor{red}{\textbf{.983}} & \textcolor{red}{\textbf{.015}} & .938 & \textcolor{red}{\textbf{.944}} & \textcolor{red}{\textbf{.972}} & \textcolor{red}{\textbf{.023}} & \textcolor{red}{\textbf{.950}} & .934 & \textcolor{red}{\textbf{.968}} & \textcolor{red}{\textbf{.028}} & .928 & \textcolor{red}{\textbf{.930}} & \textcolor{red}{\textbf{.953}} & .033 & \textcolor{red}{\textbf{.943}} & .921 & \textcolor{red}{\textbf{.958}}\\
% ... your rows here ...
\bottomrule
\end{tabular}
}
\end{table*}

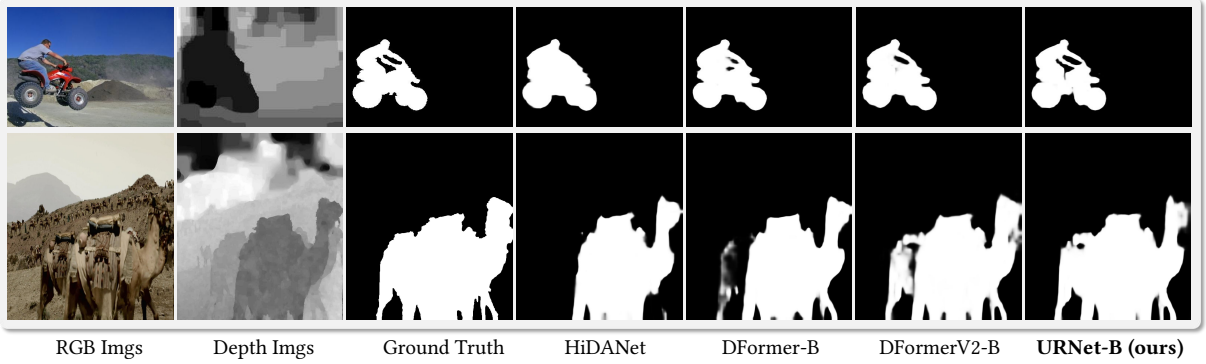
\begin{figure*}[t]
	\centering
    \begin{overpic}[width=0.9\textwidth]{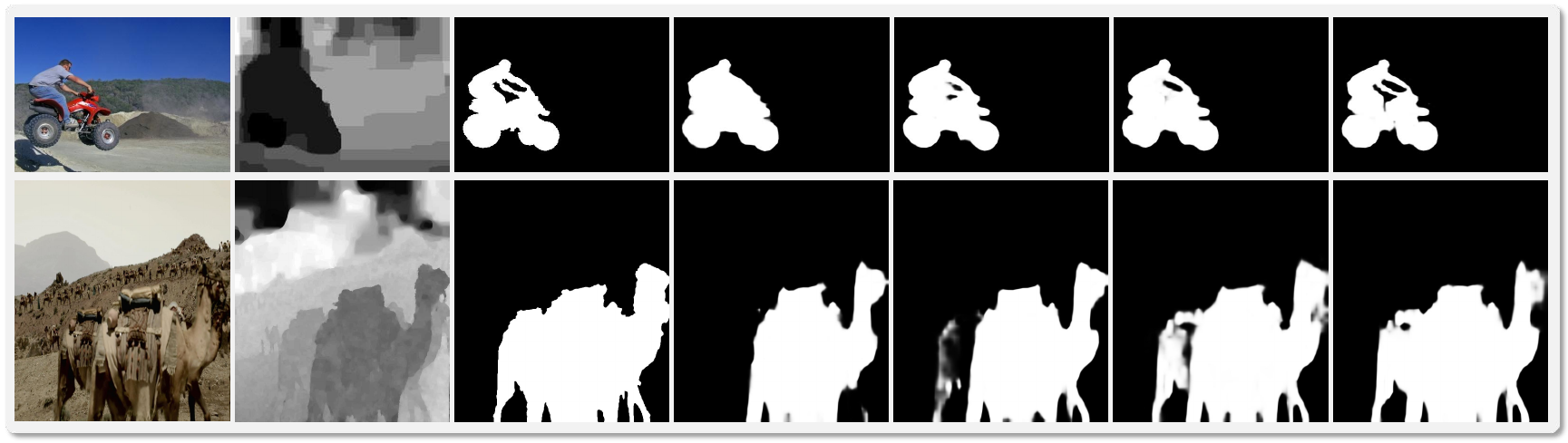}
        \put(5,-1.5){\small RGB Imgs}
        \put(18,-1.5){\small Depth Imgs}
        \put(32,-1.5){\small Ground Truth}
        \put(47,-1.5){\small HiDANet}
        \put(60,-1.5){\small DFormer-B}
        \put(73,-1.5){\small DFormerV2-B}
        \put(86,-1.5){\small \textbf{URNet-B (ours)}}
    \end{overpic}
	\caption{Qualitative Results of Salient Object Detection. From left to right are RGB images, depth images, ground truth and segmentation results produced by HiDANet~\cite{wu2023hidanet}, DFormer-B~\protect\cite{yin2024dformer}, DFormerV2-B~\protect\cite{yin2025dformerv2}, and URNet-B (ours).} 
	\label{fig:sod_vis}
    \Description{}
    \vspace{-2mm}
\end{figure*}

\textbf{Effectiveness of PMD.} We organize an ablation study of different decoder designs on the NYUDepth V2 dataset, as reported in Table~\ref{abl_pmd}. Using an MLP decoder yields the lowest performance, achieving only 51.74\% mIoU, although it maintains high inference speed. Employing HAM significantly improves accuracy to 52.78\% mIoU, but at the cost of a substantial increase in computational burden, with FLOPs rising from 13.98G to 28.40G. The proposed PMD achieves the best overall performance, reaching 53.17\% mIoU, while maintaining a favorable efficiency profile. These results demonstrate that PMD effectively enhances multi-scale feature aggregation while preserving high computational efficiency. For better comparison, the visualizations under different decoders are presented in Figure~\ref{fig:vs_dec}.

As shown in Table~\ref{abl_backbone}, which compares different encoder–decoder pairings under identical decoder settings. Across all decoders (MLP, PMD, and HAM), URNet consistently achieves higher mIoU with fewer parameters than prior encoders. In particular, URNet-B surpasses Sigma-T by +3.79\% mIoU (MLP) and +3.28\% mIoU (PMD) while reducing parameters. Similarly, URNet-L matches or slightly outperforms DFormerV2-L across HAM and PMD decoders with substantially fewer parameters, demonstrating the strong representation capacity and parameter efficiency of URNet as a plug-and-play encoder.

\subsection{Further evaluation}
\textbf{RGB-D Salient Object Detection.}
We finetune the URNet on a curated RGB-D training set and evaluate its performance on five widely used RGB-D salient object detection benchmarks. 
% Details can be found in our Appendix.
The training set contains a total of 2,195 images, including 1,485 samples from NJU2K-train and 700 samples from NLPR-train. For evaluation, we follow standard protocols and conduct experiments on DES, NLPR-test, NJU2K-test, STERE, and SIP, which contain 135, 300, 500, 1,000, and 929 images, respectively. As shown in Table~\ref{SOD}, URNet achieves consistently competitive performance across all five RGB-D salient object detection benchmarks with a highly efficient architecture. Compared with recent large-scale RGB-D SOD models, URNet-L demonstrates a more favorable accuracy–efficiency trade-off. Despite using significantly fewer parameters and FLOPs than HiDANet~\cite{wu2023hidanet} (62.0M, 22.3G vs. 130.6M, 71.5G), URNet-L achieves consistently better performance across all benchmarks.
% or comparable performance across all benchmarks. 
% Overall, URNet consistently achieves state-of-the-art or near state-of-the-art performance with substantially reduced computational cost, highlighting its practical value for large-scale RGB-D salient object detection. 
We report qualitative results on RGB-D salient object detection in Figure~\ref{fig:sod_vis}. The results indicate that URNet is able to generate more complete and accurate saliency maps, effectively suppressing background noise while preserving object integrity. These observations further validate the generalization capability of the proposed unified RGB-D representation across different dense prediction tasks.

More ablation studies and detailed analyses can be found in \href{https://github.com/Wild-Stephen/URNet}{\textcolor{purple}{Supplements}}.
\vspace{-4.5mm}

\section{Conclusion}
\label {sect:conclusion}
In this work, we presented URNet, a lightweight yet efficient RGB-D semantic segmentation framework that jointly addresses the limitations of conventional RGB-pretrained encoders and costly multi-branch fusion designs. To achieve efficient inference, we introduce a RepBlock equipped with Linear Gated Attention (LGA) for effective RGB–depth interaction, together with a lightweight Pyramid Merging Decoder (PMD) that significantly improves segmentation performance at minimal computational cost. Experiments on NYU Depth V2 and SUN-RGBD have demonstrated that URNet achieves competitive or superior performance with substantially reduced computational overhead, making it a practical solution for RGB-D scene understanding in resource-constrained settings.

\bibliographystyle{ACM-Reference-Format}
\balance
\bibliography{sample-base}

\end{document}